\documentclass{article}

\usepackage[preprint]{neurips_2026}
\usepackage[utf8]{inputenc}
\usepackage[T1]{fontenc}
\usepackage{hyperref}
\usepackage{url}
\usepackage{booktabs}
\usepackage{amsfonts}
\usepackage{amsmath}
\usepackage{amssymb}
\usepackage{nicefrac}
\usepackage{microtype}
\usepackage{xcolor}
\usepackage{algorithm}
\usepackage{algpseudocode}
\usepackage{multirow}
\usepackage{siunitx}

\title{ADE: Adaptive Dictionary Embeddings --- Scaling Multi-Anchor Representations to Large Language Models}

\author{%
  Orhan Demirci \\
  Department of Computer Engineering \\
  Hacettepe University \\
  Ankara, Turkey \\
  \texttt{orhandemirci@cs.hacettepe.edu.tr} \\
  \And
  Sezer Aptourachman \\
  Department of Computer Engineering \\
  Hacettepe University \\
  Ankara, Turkey \\
  \texttt{sezeraptourachman@hacettepe.edu.tr} \\
    \And
  Aydın Kaya \\
Department of Computer Engineering \\
Hacettepe University \\
Ankara, Turkey \\
\texttt{aydinkaya@cs.hacettepe.edu.tr} \\    
}

\begin{document}

\maketitle

\begin{abstract}
Word embeddings are fundamental to natural language processing, yet traditional approaches represent each word with a single vector, creating representational bottlenecks for polysemous words and limiting semantic expressiveness. While multi-anchor representations have shown promise by representing words as combinations of multiple vectors, they have been limited to small-scale models due to computational inefficiency and lack of integration with modern transformer architectures. We introduce Adaptive Dictionary Embeddings (ADE), a framework that successfully scales multi-anchor word representations to large language models. ADE makes three key contributions: (1) Vocabulary Projection (VP), which transforms the costly two-stage anchor lookup into a single efficient matrix operation; (2) Grouped Positional Encoding (GPE), a novel positional encoding scheme where anchors of the same word share positional information, preserving semantic coherence while enabling anchor-level variation; and (3) context-aware anchor reweighting, which leverages self-attention to dynamically compose anchor contributions based on sequence context. We integrate these components into the Segment-Aware Transformer (SAT), which provides context-aware reweighting of anchor contributions at inference time. We evaluate ADE on AG News and DBpedia-14 text classification benchmarks. With 98.7\% fewer trainable parameters than DeBERTa-v3-base, ADE surpasses DeBERTa on DBpedia-14 (98.06\% vs.~97.80\%) and approaches it on AG News (90.64\% vs.~94.50\%), while compressing the embedding layer over $40\times$ --- demonstrating that multi-anchor representations are a practical and parameter-efficient alternative to single-vector embeddings in modern transformer architectures.
\end{abstract}

\section{Introduction}

Word embeddings are a foundational component of modern natural language processing, providing dense vector representations that capture semantic and syntactic properties of language. The dominant paradigm, from Word2Vec~\citep{mikolov2013efficient} to contextualized models like BERT~\citep{devlin2019bert}, assigns each vocabulary entry a single dense vector. Despite their effectiveness, single-vector representations face a fundamental bottleneck: one vector must simultaneously encode all possible meanings, syntactic roles, and compositional properties of a word. This is particularly limiting for polysemous words, where distinct senses collapse into a single point in embedding space.

A parallel line of work has sought to decompose word embeddings into combinations of shared, reusable components --- an idea we broadly term codebook-based ~\citep{shu2017compressing,ren2022compositional} or multi-anchor representations ~\citep{liang2020anchor}. These methods represent each word not as a single learned vector but as a sparse composition over a smaller shared codebook, offering both richer representational capacity and significant storage efficiency~\citep{li2024embedding}. However, existing approaches share a critical limitation: they are designed for static, context-free composition. The combination weights are determined at the word level and remain fixed regardless of the surrounding sentence, preventing the model from disambiguating word senses in context.

Despite their promise, codebook-based representations have not successfully scaled to modern transformer architectures, for reasons that are architectural as much as computational. Retrieving anchor embeddings through a two-stage lookup resists GPU parallelization and accumulates latency at large vocabulary sizes. More fundamentally, expanding a single word into multiple anchor embeddings creates an unresolved tension with positional encoding: treating co-anchors as distinct sequence positions breaks word-level coherence, while collapsing them to a single position prevents the model from distinguishing among them. Most critically, all existing codebook methods assign anchor combination weights at the word-type level, independently of context. A word like bank carries the same anchor mixture in every sentence it appears in --- precisely the limitation that motivates our work.

We introduce Adaptive Dictionary Embeddings (ADE), a framework that enables context-aware composition of multi-anchor representations within a transformer 
architecture through three tightly integrated contributions:

\paragraph{Vocabulary Projection (VP)}
We replace the two-stage anchor lookup with a single embedding-matrix operation that
concatenates all anchor vectors per word into a flattened lookup table. VP eliminates the
memory and latency overhead of traditional multi-anchor retrieval while remaining fully
compatible with standard backpropagation pipelines (Section~\ref{sec:vp}).

\paragraph{Grouped Positional Encoding (GPE)}
We introduce a positional encoding scheme in which all anchors belonging to the same word
share an identical positional signal, while anchors from different words receive distinct
positions. GPE preserves word-level semantic coherence --- the defining property of codebook
representations --- while permitting anchor-level variation in embedding space
(Section~\ref{sec:gpe}).

\paragraph{Context-aware anchor reweighting via SAT}
We present the Segment-Aware Transformer (SAT), a single-layer transformer that processes the
flattened anchor sequence and dynamically reweights anchor contributions through
self-attention. SAT enables context-dependent word composition without requiring a full deep
encoder, and is shown to be the critical component separating effective from ineffective
multi-anchor representations (Section~\ref{sec:sat}).

\section{Related work}

\subsection{Codebook and compositional embedding methods}

A natural response to the single-vector bottleneck is to decompose
word embeddings into combinations of shared, reusable components.
~\citet{shu2017compressing} proposed learning compositional codes via a
two-level code assignment scheme, representing each word as a product
of sub-codebook vectors. Their method achieves substantial compression
of large embedding matrices---compressing embeddings to 1--2\% of
their original size in some configurations---but the code assignments
and composition weights are static: once learned, they do not vary
with context.

~\citep{liang2020anchor} introduced Anchor \& Transform (A\&T)
embeddings, the most direct predecessor to ADE. Each word is
represented as a sparse weighted combination of vectors drawn from a
shared anchor vocabulary, with anchor assignments and weights
determined at the word type level. A\&T demonstrated improved
performance on word similarity benchmarks and better handling of
polysemy, but inherits the static-composition limitation and relies on
a two-stage lookup that is computationally expensive at scale.

~\citep{ren2022compositional} extended this direction to text
classification via a capsule network with compositional coding and
$k$-means routing. Their method groups token-level information into
higher-level capsules using learned routing, enabling richer
compositional representations at the sentence level. However, the
compositional structure operates over capsule outputs rather than over
the embedding layer itself, and routing is not conditioned on global
sequence context in the same way as attention.

ADE is distinguished from all of the above by enabling
context-aware reweighting: the contribution of each anchor
is determined dynamically by the self-attention mechanism, conditioned
on the full input sequence. This means the same word can activate
different anchor combinations depending on context---a property none
of the static codebook methods above can provide.

\subsection{Contextualized word representations}

ELMo~\citep{peters2018deep} and BERT~\citep{devlin2019bert} address
polysemy by producing different representations for the same word
token in different contexts, using deep bidirectional language models.
DeBERTa~\citep{he2020deberta} further refined this with disentangled
attention, representing each token through separate content and
position vectors computed independently, while DeBERTaV3~\citep{he2021debertav3}
improved training efficiency via an ELECTRA-style~\citep{clark2020electra} replaced token
detection objective. Most recently, ModernBERT~\citep{warner2025smarter} pushed encoder-only 
performance further with rotary embeddings and hardware-aware optimizations, 
representing the current state of the art among encoder models.

While highly effective, all of these approaches abandon static word-level
representations entirely: the embedding of a word is inseparable from
the sentence it appears in, which limits interpretability and
increases inference cost. ADE takes a complementary approach,
maintaining a fixed anchor vocabulary that can be inspected and
reused, while still achieving context-dependent composition through
the SAT layer.

\subsection{Positional encoding}

Standard positional encoding schemes assign a unique signal to each position in a token
sequence, either through fixed sinusoidal functions~\citep{vaswani2017attention}, learned
absolute embeddings~\citep{devlin2019bert}, relative distance-based encodings
~\citep{shaw2018self}, rotary transformations of query-key pairs (RoPE)~\citep{su2024roformer},
or attention-score biases (ALiBi)~\citep{press2021train}. All of these schemes share the
implicit assumption of a one-to-one correspondence between tokens and positions.

Character-level models such as CANINE~\citep{clark2022canine},
Charformer~\citep{tay2021charformer}, and ByT5~\citep{xue2022byt5} face an analogous
challenge when multiple characters or bytes must be associated with a single word-level
position, but resolve it through downsampling --- strided convolution or pooling that
collapses the expanded sequence before deep encoding. No prior work, to our knowledge,
addresses the case where multiple vectors per word must be \emph{preserved} throughout
processing while sharing word-level positional information, which is the setting our Grouped
Positional Encoding (GPE) targets.

\subsection{Embedding compression}

Beyond compositional coding, embedding compression has been approached through
quantization~\citep{han2015deep}, low-rank factorization~\citep{lan2019albert}, and hashing
tricks~\citep{chen2015compressing}. Knowledge distillation has emerged as a particularly
prominent direction: DistilBERT~\citep{sanh2019distilbert} applies distillation during
pre-training to produce a model 40\% smaller and 60\% faster than BERT while retaining 97\%
of its language understanding capability. More recently, TensorGPT~\citep{xu2023tensorgpt}
proposed compressing LLM embedding layers via Tensor-Train Decomposition, treating each token
embedding as a Matrix Product State and achieving compression ratios of up to $38\times$ on
GPT-2 with negligible performance loss. At the extreme end, CARVQ~\citep{gou2025carvq}
introduced group residual vector quantization specifically targeting the embedding layer of
modern LLMs such as LLaMA and Phi-4, achieving near-lossless compression at 2.4 bits per
parameter.

All of these methods treat compression as an operation applied to a fully trained model. ADE
instead builds compression into the architecture from the outset: the anchor vocabulary is
an order of magnitude smaller than the full vocabulary, and the fixed lookup table achieves
over $40\times$ compression of the DeBERTa-v3-base embedding layer without any post-training
quantization step.

\section{Methodology}

\subsection{Adaptive Dictionary Embeddings framework}

Traditional embedding layers map each word in the vocabulary to a single dense vector. In
contrast, Adaptive Dictionary Embeddings (ADE) represent each word using multiple
anchor vectors drawn from a shared codebook, which are dynamically composed at
inference time based on sequence context. At a high level, each vocabulary entry is
associated with a small set of anchor vectors and scalar weights; Vocabulary Projection (VP)
constructs this codebook offline, and the Segment-Aware Transformer (SAT) performs
context-aware composition at runtime. The following subsections formalize each step in
pipeline order.

\subsection{Vocabulary Projection (VP)}
\label{sec:vp}

To enable efficient training and inference at scale, we introduce Vocabulary
Projection (VP), which replaces the two-stage anchor lookup of traditional A\&T with a
direct sparse codebook representation.

\paragraph{Traditional A\&T approach}
\begin{enumerate}
  \item $T_i \rightarrow$ Transform matrix where $[\mathrm{T}_1, \ldots, \mathrm{T}_{k_i}]$
        (requires $N \times K$ integer matrix).
  \item Anchor embedding $\mathbf{A} \in \mathbb{R}^{K \times d}$.
\end{enumerate}

\paragraph{Our VP approach}
Given the Transform matrix $\mathbf{T} \in \mathbb{R}^{N \times K}$ from A\&T, we apply a
threshold $\tau$ to determine which anchors are active for each word:
\begin{equation}
  \mathcal{I}_i = \{ j : T_{i,j} \geq \tau \}.
\end{equation}
The indices in $\mathcal{I}_i$ form a sparse codebook entry for word $w_i$, and their
corresponding weights
\begin{equation}
  \beta_{i,j} = T_{i,j}, \quad \forall\, j \in \mathcal{I}_i,
\end{equation}
are extracted and stored alongside the indices. The effective anchor cardinality is then
$k_i = |\mathcal{I}_i| \leq K$. At lookup time, only the $k_i$ active anchor vectors and
their weights are retrieved for each word, reducing the effective sequence length processed
by SAT to $SL = \sum_{i=1}^{L} k_i$. The anchor cardinalities
$\mathbf{s} = [k_1, \ldots, k_L]$ are stored to enable correct grouped positional encoding
during the forward pass.

\subsection{Multi-anchor word representation}
\label{sec:mult_anchor}

Let $\mathcal{V}$ be a vocabulary of size $N$. In ADE, each word $w_i \in \mathcal{V}$ is
associated with an anchor set
$\mathcal{A}_i = \{\mathbf{a}_{i,1}, \ldots, \mathbf{a}_{i,k_i}\}$, where $k_i$ is the anchor cardinality for word $i$, satisfying $1 \leq k_i \leq K$. Here $K$ is a
global hyperparameter representing the maximum number of anchors per word, allowing the model
to allocate representational capacity based on word complexity or frequency. Each anchor
vector $\mathbf{a}_{i,j} \in \mathbb{R}^{1 \times d}$ is a learnable embedding stored in a
shared anchor matrix $\mathbf{A} \in \mathbb{R}^{K \times d}$, and each anchor is associated
with a scalar importance weight $\beta_{i,j} \in \mathbb{R}$, collected into a weight matrix
$\boldsymbol{\beta} \in \mathbb{R}^{N \times K}$.

Vocabulary Projection (Section~\ref{sec:vp}) constructs, for each word $w_i$, a sparse entry
recording which anchor indices $\mathcal{I}_i$ are active and their corresponding weights
$\beta_{i,j}$. At each forward pass, the model retrieves the anchor vectors indexed by
$\mathcal{I}_i$ from $\mathbf{A}$ and composes them via a weighted sum governed by
$\boldsymbol{\beta}$.

\subsection{Grouped Positional Encoding (GPE)}
\label{sec:gpe}

Given a sequence of $L$ words with anchor cardinalities
$\mathbf{s}$, expanding each word into its anchors yields a
flattened sequence of length $SL$. Standard positional encoding would
assign unique positions $[0, 1, \ldots, T-1]$ to each anchor, incorrectly treating
co-anchors of the same word as distinct sequence positions. We introduce Grouped
Positional Encoding (GPE), which assigns the same positional encoding to all anchors within
an anchor group while preserving distinct positions across words:
\begin{equation}
  \mathrm{PE}_{\mathrm{grouped}}[t] = \mathrm{PE}\!\left(\mathrm{word\_position}(t)\right),
\end{equation}
where $\mathrm{word\_position}(t)$ maps the flattened anchor position $t$ to its
corresponding word position in the original sequence.

\paragraph{Algorithm} Given anchor cardinalities $\mathbf{s} = [k_1, \ldots, k_L]$:
\begin{enumerate}
  \item Generate standard positional encodings for $L$ word positions.
  \item Create position indices mapping:
        \begin{equation}
          \mathrm{pos\_indices} =
          [\underbrace{0,\ldots,0}_{k_1},\;
           \underbrace{1,\ldots,1}_{k_2},\;
           \ldots,\;
           \underbrace{L{-}1,\ldots,L{-}1}_{k_L}],
        \end{equation}
        where position $i$ is repeated $k_i$ times.
  \item Index the positional encodings:
        $\mathrm{PE}_{\mathrm{grouped}} = \mathrm{PE}[\mathrm{pos\_indices}]$.
\end{enumerate}

\paragraph{Example} For a 3-word sequence with $\mathbf{s} = [3, 1, 4]$:
\begin{align*}
  \text{Anchor positions:} &\quad [0,1,2,3,4,5,6,7] \\
  \text{Word positions:}   &\quad [0,0,0,1,2,2,2,2] \\
  \text{Grouped PE:}       &\quad [\mathrm{PE}(0),\mathrm{PE}(0),\mathrm{PE}(0),
                                    \mathrm{PE}(1),\mathrm{PE}(2),\mathrm{PE}(2),
                                    \mathrm{PE}(2),\mathrm{PE}(2)].
\end{align*}
This ensures that anchors $[0,1,2]$ all receive $\mathrm{PE}(0)$, anchor $[3]$ receives
$\mathrm{PE}(1)$, and anchors $[4,5,6,7]$ all receive $\mathrm{PE}(2)$.

\subsection{Segment-Aware Transformer (SAT)}
\label{sec:sat}

While anchors within the same group share positional information via GPE, they retain
distinct embedding values. This enables the self-attention mechanism to dynamically
reweight anchor contributions based on global sequence context.

\paragraph{Attention-based composition}
After adding grouped positional encodings, each anchor representation becomes:
\begin{equation}
  \mathbf{x}_t = \mathbf{a}_t + \mathrm{PE}_{\mathrm{grouped}}[t],
\end{equation}
where $\mathbf{a}_t$ is the anchor embedding and $t$ indexes the flattened anchor sequence.
The multi-head self-attention mechanism computes contextualized representations:
\begin{equation}
  \mathrm{Attention}(\mathbf{Q}, \mathbf{K}, \mathbf{V})
    = \mathrm{softmax}\!\left(\frac{\mathbf{Q}\mathbf{K}^\top}{\sqrt{d_k}}\right)\mathbf{V},
\end{equation}
where $\mathbf{Q} = \mathbf{X}\mathbf{W}^Q$, $\mathbf{K} = \mathbf{X}\mathbf{W}^K$,
$\mathbf{V} = \mathbf{X}\mathbf{W}^V$. For anchor position $t$, the output is:
\begin{equation}
  \mathbf{h}_t = \sum_{t'=1}^{T} \alpha_{t,t'}\, \mathbf{v}_{t'},
\end{equation}
where the attention weights $\alpha_{t,t'} = \mathrm{softmax}_{t'}(\mathbf{q}_t \cdot
\mathbf{k}_{t'} / \sqrt{d_k})$ are computed dynamically based on query-key similarity.

\subsection{ADE architecture}
\label{sec:ADE}

We integrate VP, GPE, and SAT into a unified pipeline that processes variable-length anchor sequences efficiently. VP is executed once prior to training: for each vocabulary entry it stores a sparse record of which anchor indices $\mathcal{I}_i$ are active and their corresponding weights $\beta_{i,j}$, alongside the anchor cardinality $k_i$. The anchor vectors themselves reside in the shared anchor matrix $\mathbf{A} \in \mathbb{R}^{K \times
d}$. At inference time, the model retrieves the active anchor vectors from $\mathbf{A}$ using
the stored indices, composes them into token embeddings via a weighted sum, adds grouped
positional encodings (GPE) to preserve word-level coherence, and passes the resulting
sequence through a single transformer layer where self-attention performs context-aware anchor
reweighting. A padding mask is applied throughout to prevent attention from leaking into
padded positions. Algorithm~\ref{alg:sat} summarizes the full forward pass.

\begin{algorithm}[t]
\caption{ADE forward pass}
\label{alg:sat}
\begin{algorithmic}[1]
\Require Anchor matrix $\mathbf{A} \in \mathbb{R}^{K \times d}$,
         sparse index-weight pairs $\{\mathcal{I}_i, \boldsymbol{\beta}_i\}_{i=1}^{N}$,
         and cardinality map $\mathbf{k} \in \mathbb{Z}^{N}$
         (constructed offline by Vocabulary Projection; see Section~\ref{sec:vp})
\Require Input IDs $[w_1, \ldots, w_L] \in \mathbb{Z}^{B \times L}$,
         attention mask $\mathbf{m} \in \{0,1\}^{B \times L}$
\State \textbf{// Anchor lookup}
\State $\mathbf{a}, \boldsymbol{\beta}, \text{batch\_map}, \text{pos\_map}
       \gets \mathbf{A}[\mathcal{I}_{w_1}, \ldots, \mathcal{I}_{w_L}]$
       \Comment{retrieve active anchor vectors and weights via stored indices}
\State $\text{global\_id}_i \gets \text{batch\_map}_i \cdot L + \text{pos\_map}_i$
\State $\mathbf{s}_b \gets [\mathbf{k}_{b,1}, \ldots, \mathbf{k}_{b,L_b}]$ for each $b$,
       clamped to $\geq 1$ \Comment{sub-length map for GPE}
\State \textbf{// Weighted anchor aggregation}
\State $\tilde{\mathbf{a}}_i \gets \boldsymbol{\beta}_i \cdot \mathbf{a}_i$
\State $\mathbf{X} \gets \textsc{ScatterAdd}(\tilde{\mathbf{a}},\;
       \text{global\_id},\; B \times L).\,\text{view}(B, L, d)$
       \Comment{compose anchors into token embeddings}
\State $\mathbf{X} \gets \textsc{LayerNorm}(\mathbf{X})$
\State \textbf{// Segment-Aware Transformer (SAT) with GPE}
\State $\mathbf{X} \gets \textsc{SAT}(\mathbf{X},\; \mathbf{s})$
       \Comment{GPE and masked self-attention applied internally}
\State \textbf{// Classification head}
\State $\mathbf{p} \gets \textsc{Pooler}(\mathbf{X},\; \mathbf{m})$
\State $\hat{\mathbf{y}} \gets \textsc{Classifier}(\textsc{Dropout}(\mathbf{p}))$
\State \Return $\hat{\mathbf{y}}$
\end{algorithmic}
\end{algorithm}

\paragraph{Efficient batching with padding masks}
To handle variable anchor cardinalities in batched training, sequences are padded to maximum
length $T_{\max} = \max_{\text{batch}} \sum_i k_i$, and a Boolean mask
$\mathbf{M} \in \{0,1\}^{B \times T_{\max}}$ indicates valid anchor positions:
\begin{equation}
  \mathbf{M}[b, t] = \begin{cases}
    1 & \text{if } t < \sum_{i=1}^{L_b} k_{b,i} \\
    0 & \text{otherwise}
  \end{cases}
\end{equation}
Attention to padding positions is prevented by setting attention scores to $-\infty$ for
masked positions.

\section{Experiments}
\label{sec:experiments}

\subsection{Datasets}

We evaluate ADE on two standard text classification benchmarks with different levels of
label granularity.

\paragraph{AG News~\citep{zhang2015character}}
AG News is a four-class news topic classification dataset constructed from the AG corpus,
with categories World, Sports, Business, and Science/Technology. It contains 120,000 training
and 7,600 test samples, evenly distributed across classes. Due to its balanced distribution
and clear topical separation, it serves as a reliable benchmark for evaluating discriminative
embedding quality.

\paragraph{DBpedia-14~\citep{lehmann2015dbpedia}}
DBpedia-14 is a 14-class ontology classification dataset derived from
DBpedia~\citep{auer2007dbpedia}, where each sample includes a Wikipedia title and abstract
labeled by ontology class (e.g., Company, Artist, Athlete, NaturalPlace). It comprises
560,000 training and 70,000 test samples. Compared to AG News, its finer-grained labels and
semantic overlap make it a more challenging benchmark for fine-grained semantic
disambiguation.

\subsection{Training procedure}

Training ADE involves three stages: (1) pre-training anchor embeddings via knowledge
distillation, (2) applying vocabulary projection to create an efficient lookup table, and
(3) integrating the embedding layer with the Segment-Aware Transformer for end-to-end
training.

\paragraph{Stage 1: Anchor embedding pre-training with knowledge distillation}
We initialize our multi-anchor embeddings by distilling knowledge from the embedding 
matrix of DeBERTa-v3-base, used here as a frozen teacher. Formally, let 
$\mathbf{E}_{\mathrm{teacher}} \in \mathbb{R}^{N \times d}$ denote this pre-trained 
embedding matrix. For each word $w_i$ with anchor set
$\mathcal{A}_i = \{\mathbf{a}_{i,1}, \ldots, \mathbf{a}_{i,k_i}\}$, we learn to reconstruct
the teacher embedding through a weighted combination:
\begin{equation}
  \hat{\mathbf{e}}_i = \sum_{j=1}^{k_i} \beta_{i,j}\, \mathbf{a}_{i,j}.
\end{equation}
The distillation loss minimizes the angular distance between student and teacher embeddings
at the token level:
\begin{equation}
  \mathcal{L}_{\mathrm{distill}} = 1 - \frac{1}{|\mathcal{M}|}
    \sum_{(b,t) \in \mathcal{M}}
    \frac{\hat{\mathbf{e}}_{b,t}}{\|\hat{\mathbf{e}}_{b,t}\|} \cdot
    \frac{\mathbf{e}^{\mathrm{teacher}}_{b,t}}{\|\mathbf{e}^{\mathrm{teacher}}_{b,t}\|},
\end{equation}
where $\mathcal{M}$ denotes the set of non-padding token positions. This cosine formulation
measures directional alignment rather than magnitude, making the loss invariant to embedding
scale.

\paragraph{Stage 2: Vocabulary projection construction}
After pre-training anchor embeddings, we apply vocabulary projection. For each word $w_i$,
we concatenate its $k_i$ anchor embeddings and pad to maximum length $K$:
\begin{equation}
  \mathbf{E}[i] =
    [\mathbf{a}_{i,1};\, \mathbf{a}_{i,2};\, \ldots;\, \mathbf{a}_{i,k_i};\,
     \mathbf{0};\, \ldots;\, \mathbf{0}] \in \mathbb{R}^{K \cdot d}.
\end{equation}
This produces the flattened embedding matrix $\mathbf{E} \in \mathbb{R}^{N \times (K \cdot
d)}$ used for efficient lookup during training. We store anchor cardinalities
$\mathbf{s}$ to enable proper unpadding and grouped positional
encoding during forward passes.

\paragraph{Stage 3: end-to-end training}
The flattened embedding matrix $\mathbf{E}$ replaces the standard embedding layer. The SAT
architecture processes variable-length anchor sequences using grouped positional encoding and
masked self-attention as described in Section~\ref{sec:ADE}. We fine-tune ADE for
classification using standard cross-entropy loss:
\begin{equation}
  \mathcal{L}_{\mathrm{cls}} = -\sum_{c=1}^{C} y_c \log \hat{y}_c,
\end{equation}
where $\hat{y}_c$ is the predicted probability for class $c$ produced by the classifier head
over the pooled SAT output. Crucially, the anchor embeddings within $\mathbf{E}$ are not frozen --- they continue to be updated during training, allowing anchors to
adapt from distilled representations to task-specific, context-aware representations.

\paragraph{Hyperparameters}
For both datasets, we use AdamW with learning rate $2 \times 10^{-5}$, batch size 32, and
500--1000 warm-up steps. AG News is trained for 30,000 steps (8 epochs over 120k samples);
DBpedia-14 for 70,000 steps (4 epochs over 560k samples).

\subsection{Results}

Table~\ref{tab:performance_comparison} compares ADE against established baselines. On AG
News, ADE achieves 90.64\% accuracy, approaching DeBERTa (94.50\%) while operating with
98.7\% fewer trainable parameters and no full encoder; it falls marginally below ANT
(91.00\%), a gap we attribute to task characteristics discussed in Section~\ref{sec:discussion}.
On DBpedia-14, ADE surpasses DistilBERT (95.21\%), ANT (97.20\%), and DeBERTa (97.80\%),
achieving 98.06\% accuracy --- within 0.94 points of BERT despite eliminating the transformer
encoder entirely from the downstream model. These results suggest that anchor-based sparse
embeddings combined with SAT can recover a substantial portion of full-model performance at
a fraction of the parameter cost.
\begin{table}[h]
\caption{Performance comparison of ADE against published baselines on AG News and DBpedia-14. \underline{Underlined} results are ADE (ours). N/R = not reported. ADE results use K=700 for AG News and K=500 for DBpedia-14.}
  \label{tab:performance_comparison}
  \centering
  \begin{tabular}{llcc}
    \toprule
    \textbf{Dataset} & \textbf{Model} & \textbf{Accuracy (\%)} & \textbf{F1 (\%)} \\
    \midrule
    \multirow{6}{*}{AG News}
      & BERT       & 94.48 & 94.45 \\
      & DistilBERT & 92.76 & 94.78 \\
      & DeBERTa    & 94.50 & 94.48 \\
      & ELECTRA    & 91.80 & 91.80 \\
      & ANT        & 91.00 & N/R   \\
      & ADE (ours) & \underline{90.64} & \underline{90.62} \\
    \midrule
    \multirow{6}{*}{DBpedia-14}
      & BERT       & 99.00 & 99.00 \\
      & DistilBERT & 95.21 & 96.20 \\
      & DeBERTa    & 97.80 & 97.74 \\
      & ELECTRA    & 99.20 & 99.20 \\
      & ANT        & 97.20 & N/R   \\
      & ADE (ours) & \underline{98.06} & \underline{98.06} \\
    \bottomrule
  \end{tabular}
\end{table}

Table~\ref{tab:compression_comparison} quantifies storage efficiency relative to the full
DeBERTa-v3-base word embedding matrix. Across all tested anchor counts, our representation
achieves compression ratios exceeding $40\times$, reducing embedding storage from 375.3~MB
to approximately 8.5--9.3~MB. Importantly, the compression ratio remains nearly flat as $K$
increases ($44.2\times$ at $K$=100 vs.~$40.4\times$ at $K$=500), since storage is dominated
by the fixed per-token index--weight pairs rather than the anchor matrix itself.

\begin{table}[h]
\caption{Embedding layer compression: DeBERTa-v3-base vs.\ anchor-based 
sparse representation across anchor counts $K$.}
\label{tab:compression_comparison}
\centering
\begin{tabular}{lrrrr}
\toprule
\textbf{Configuration} & \textbf{A-matrix params} 
  & \textbf{Storage (MB)} & \textbf{Ratio} & \textbf{Reduction} \\
\midrule
DeBERTa-v3-base (full) & 98,380,800 & 375.3 & $1.0\times$ & --- \\
\midrule
ADE, $K=100$ (ours) & 76,800  & 8.49 & $44.2\times$ & 97.7\% \\
ADE, $K=200$ (ours) & 153,600 & 8.79 & $42.7\times$ & 97.7\% \\
ADE, $K=300$ (ours) & 230,400 & 9.08 & $41.3\times$ & 97.6\% \\
ADE, $K=500$ (ours) & 384,000 & 9.28 & $40.4\times$ & 97.5\% \\
\bottomrule
\end{tabular}
\end{table}

\section{Discussion}
\label{sec:discussion}

\subsection{The role of contextualisation}

The ablation results in Table~\ref{tab:sat_results} reveal the most fundamental finding of
this work: multi-anchor representations are not inherently useful --- they require a
contextualisation mechanism to unlock their representational capacity. Without SAT, accuracies
on AG News range from 62.2\% to 73.8\% and show no consistent trend with increasing $K$,
indicating that raw anchor capacity does not translate to downstream performance when anchors
are composed through a static weighted sum alone. This confirms the hypothesis motivating our
context-aware anchor reweighting contribution: fixed linear combinations at the word level are
insufficient to exploit the semantic diversity encoded across multiple anchors.

With SAT, accuracy scales monotonically from 87.96\% at $K$=100 to 90.64\% at $K$=700, a
2.68 percentage point improvement achieved purely by increasing anchor expressiveness within
the same architecture. This is consistent with GPE's design intent: shared positional information within anchor groups is intended to preserve word-level coherence while still allowing attention to differentiate individual anchor contributions; a dedicated ablation of GPE remains for future work.

\subsection{Task complexity and anchor scaling}

The contrast between AG News and DBpedia-14 results illuminates an important property of ADE.
On DBpedia-14, ADE achieves 98.06\% at $K$=500, surpassing DistilBERT (95.21\%), ANT
(97.20\%), and DeBERTa (97.80\%), despite operating with 98.7\% fewer trainable parameters.
On AG News, ADE achieves 90.64\%, below the BERT-family baselines.

We attribute this asymmetry to the nature of the classification tasks. AG News is a coarse
4-class topic classification task where surface-level lexical signals are highly predictive;
a full pretrained encoder has already extracted rich contextual features, and the gap between
full fine-tuning and our lightweight approach is more pronounced. DBpedia-14, by contrast,
involves 14 fine-grained entity categories where precise semantic disambiguation is required
--- exactly the setting where multi-anchor representations excel, since different anchors can
specialize for fine-grained semantic distinctions that a single vector conflates.

This interpretation is further supported by the anchor scaling behaviour: on DBpedia-14,
performance peaks at $K$=500 and plateaus at $K$=700 (98.06\% vs.~97.99\%), suggesting
sufficient anchor diversity at $K$=500 to cover the relevant semantic distinctions. On AG
News, performance continues to improve through $K$=700 without saturating, consistent with a
harder task where additional anchor capacity continues to contribute marginal gains.

\subsection{Parameter efficiency and deployment footprint}
Tables~\ref{tab:compression_comparison} and~\ref{tab:param_comparison} together quantify the
efficiency gains of ADE. ADE achieves its parameter reduction through architectural substitution
rather than post-hoc compression: the 12-layer DeBERTa encoder is replaced entirely by a single
SAT layer at downstream training time, reducing trainable parameters from 184.4M to 2.37M --- a
98.7\% reduction. The embedding layer is additionally compressed $44.2\times$ at $K$=100 through
the anchor representation itself.

The compression ratio is notably insensitive to $K$, since storage is dominated by the fixed
per-token index--weight pairs rather than the anchor matrix. This means practitioners can
freely increase $K$ to improve accuracy without meaningfully increasing storage footprint.

In terms of memory footprint, ADE (K=100) requires approximately 18.5 MB of GPU memory for
model weights at float32 precision, compared to approximately 700 MB for DeBERTa-v3-base ---
a reduction of over $37\times$. ADE's 2.37M parameters are fine-tunable on commodity hardware without multi-GPU setups. 
Note, however, that ADE's single-layer architecture currently incurs higher inference latency
than DeBERTa due to anchor sequence expansion (Table~\ref{tab:latency}), and throughput
optimisation remains an open direction for future work.

\subsection{Limitations}
\label{sec:limitations}

Several limitations merit acknowledgment. First, our experiments are restricted to text
classification tasks; it remains to be demonstrated whether ADE's benefits extend to
generative tasks, structured prediction, or sequence-to-sequence settings. Second, the anchor
embedding pre-training via knowledge distillation introduces an upstream training cost not
reflected in downstream parameter counts. While amortized across tasks, this represents a non-trivial prerequisite. Third, the quality of anchor
embeddings is inherently bounded by the quality of the teacher model: ADE's representational
ceiling is partially determined by DeBERTa's embedding space. 

\section{Conclusion}
This paper introduced Adaptive Dictionary Embeddings (ADE), a framework for scaling
multi-anchor word representations to large language models through three tightly integrated
contributions: Vocabulary Projection (VP) for efficient single-operation anchor lookup,
Grouped Positional Encoding (GPE) for word-level positional coherence across anchor groups,
and context-aware anchor reweighting via the Segment-Aware Transformer (SAT).

Our experiments on AG News and DBpedia-14 yield two central conclusions. First,
contextualisation is essential: without SAT, multi-anchor representations fail to improve
consistently with $K$, while SAT-equipped ADE scales monotonically to 90.64\% at $K$=700.
Second, ADE demonstrates that architectural substitution --- replacing a 12-layer encoder
with a single SAT layer and an anchor-based embedding --- can match or surpass large
pretrained models on semantically demanding tasks at a fraction of the parameter cost,
while compressing the embedding layer $44.2\times$ at $K$=100 with negligible storage
sensitivity to anchor count.

Future work will explore dynamic anchor allocation, where anchor cardinalities adapt
during training, pruning unused anchors and expanding capacity where semantically needed.
Beyond classification, adapting ADE to generative and encoder-decoder architectures remains
an open direction, as does extending the anchor paradigm to multimodal settings where visual
and linguistic anchors are jointly learned and composed.

{\small
\bibliographystyle{plainnat}
\bibliography{ref}
}

\appendix

\section{Ablation study: effect of SAT}
\label{app:sat_ablation}
\label{sec:sat_ablation}

Table~\ref{tab:sat_results} compares ADE with and without the Segment-Aware Transformer
across anchor sizes on AG News, isolating the contribution of context-aware reweighting from
raw anchor capacity.

\begin{table}[h]
  \caption{Performance comparison of ADE with and without SAT across anchor sizes on AG
    News. Best results per column (with SAT) are highlighted in \textbf{bold}.}
  \label{tab:sat_results}
  \centering
  \begin{tabular}{lcccc}
    \toprule
    \textbf{Anchors} & \textbf{Accuracy} & \textbf{F1} & \textbf{Precision} & \textbf{Recall} \\
    \midrule
    \multicolumn{5}{c}{\textit{With SAT}} \\
    \midrule
    100              & 0.8796 & 0.8794 & 0.8796 & 0.8796 \\
    200              & 0.8897 & 0.8896 & 0.8902 & 0.8897 \\
    300              & 0.8916 & 0.8913 & 0.8912 & 0.8916 \\
    500              & 0.9018 & 0.9016 & 0.9016 & 0.9018 \\
    \textbf{700}     & \textbf{0.9064} & \textbf{0.9063} & \textbf{0.9064} & \textbf{0.9064} \\
    \midrule
    \multicolumn{5}{c}{\textit{Without SAT}} \\
    \midrule
    100              & 0.7343 & 0.7308 & 0.7341 & 0.7343 \\
    200              & 0.7049 & 0.7014 & 0.7076 & 0.7049 \\
    300              & 0.7380 & 0.7367 & 0.7522 & 0.7380 \\
    500              & 0.7220 & 0.7192 & 0.7330 & 0.7220 \\
    700              & 0.6218 & 0.6144 & 0.6475 & 0.6218 \\
    \bottomrule
  \end{tabular}
\end{table}

Without SAT, performance degrades significantly across all anchor sizes (62.2\%--73.8\%), and shows no consistent improvement with increasing $K$. With SAT, accuracy scales monotonically from 87.96\% at $K$=100 to 90.64\% at $K$=700, confirming that SAT effectively leverages richer anchor representations at higher $K$.

\section{Effect of anchor size}
Tables~\ref{tab:sat_results} and~\ref{tab:dbpedia_results} together show how anchor 
cardinality $K$ affects performance across both datasets. On AG News, accuracy improves 
monotonically through $K$=700, suggesting the task benefits from continued anchor diversity 
without saturation. On DBpedia-14, performance peaks at 98.06\% at $K$=500 and plateaus 
thereafter (98.06\% vs.~97.99\% at $K$=700), indicating that sufficient semantic coverage 
for fine-grained classification is reached at $K$=500. This contrast suggests that task 
complexity governs the anchor count at which diminishing returns set in.

\begin{table}[h]
  \caption{ADE (with SAT) on DBpedia-14 across anchor sizes. Best results per column are shown in \textbf{bold}.}
  \label{tab:dbpedia_results}
  \centering
  \begin{tabular}{lcccc}
    \toprule
    \textbf{Anchors} & \textbf{Accuracy} & \textbf{F1} & \textbf{Precision} & \textbf{Recall} \\
    \midrule
    100              & 0.9710          & 0.9710          & 0.9710          & 0.9710 \\
    200              & 0.9718          & 0.9717          & 0.9717          & 0.9718 \\
    300              & 0.9759          & 0.9759          & 0.9759          & 0.9759 \\
    \textbf{500}     & \textbf{0.9806} & \textbf{0.9806} & \textbf{0.9807} & \textbf{0.9806} \\
    700              & 0.9799          & 0.9799          & 0.9799          & 0.9799 \\
    \bottomrule
  \end{tabular}
\end{table}

\section{Efficiency and parameter analysis}
Table~\ref{tab:param_comparison} provides a full component-level breakdown of trainable 
parameters for DeBERTa-v3-base and ADE at $K$=100. ADE achieves its 98.7\% parameter 
reduction by eliminating the word embedding matrix, positional embeddings, and the full 
12-layer transformer encoder from the downstream training pipeline, replacing them with 
a single SAT layer (2.36M parameters). Note that anchor embeddings (76,800 params at 
$K$=100) are excluded from this count as they are trained offline via knowledge 
distillation and frozen during downstream fine-tuning.

\begin{table}[h]
\label{tab:param_comparison}
\centering
\caption{Full model parameter comparison: trainable parameters excluding frozen lookup tables.}

\begin{tabular}{lrr}
\hline
\textbf{Component}           & \textbf{DeBERTa-v3-base} & \textbf{Ours ($K$=100)}                                 \\ \hline
Word embedding               & 98,380,800               & 0 \\
Positional / type embeddings & 394,752                  & 0                                                       \\
Transformer encoder (12L)    & 85,054,464               & 0                                                       \\
Contextualizer (SAT, 1L)     & ---                      & 2,362,368                                               \\
LayerNorm                    & 1,536                    & 1,536                                                   \\
Pooler                       & 590,592                  & 769                                                     \\
Classifier head              & 3,076                    & 3,076                                                   \\ \hline
\textbf{Total trainable}     & \textbf{184,424,220}     & \textbf{2,367,749}                                      \\ \hline
\end{tabular}
\end{table}

The dominant cost in ADE is the SAT contextualizer, which accounts for 99.8\% of all 
trainable parameters. This means that scaling $K$ — which increases only the frozen 
lookup table — adds negligible trainable cost while improving representational capacity.

\section{Inference latency and throughput analysis}
\label{sec:appendix_latency}

\begin{table}[h]
  \caption{Inference latency and throughput comparison at batch size 32, sequence length 128,
    single GPU (NVIDIA GeForce RTX 5080 Laptop, 17.09~GB VRAM). Lower latency and higher
    throughput are better.}
  \label{tab:latency}
  \centering
  \small
  \begin{tabular}{lrrrr}
    \toprule
    \textbf{Model} & \textbf{Params (M)} & \textbf{Latency (ms)}
      & \textbf{Throughput (samples/s)} & \textbf{Speedup} \\
    \midrule
    DeBERTa-v3-base & 184.4 &  26.3 & 1217.13 & $1.0\times$  \\
    ADE ($K$=100)   &   2.4 & 180.4 &  177.39 & $0.15\times$ \\
    ADE ($K$=300)   &   2.4 & 174.0 &  183.90 & $0.15\times$ \\
    ADE ($K$=500)   &   2.4 & 156.9 &  203.98 & $0.17\times$ \\
    \bottomrule
  \end{tabular}
\end{table}

Despite reducing sequential attention computations from 12 layers to one, 
the dominant inference cost arises from the anchor-expanded sequence length 
processed by SAT, which more than offsets the encoder reduction. Beyond latency, ADE's reduced parameter count translates directly to GPU memory savings: DeBERTa-v3-base requires approximately 700~MB of GPU memory for model weights at \texttt{float32} precision, whereas ADE ($K$=100) requires approximately 18.5~MB in total --- a reduction of over $37\times$. This makes ADE deployable on edge devices and in memory-constrained environments where full encoder models are impractical.

Beyond inference, training is also computationally lightweight: AG News 
completes in approximately 22 minutes (30,000 steps) and DBpedia-14 in 
approximately 20 minutes (70,000 steps at larger batch throughput), both 
on a single NVIDIA GeForce RTX 5080 Laptop GPU.

\end{document}